\pdfoutput=1

\documentclass[preprint]{elsarticle}

\usepackage{url} 
\usepackage{natbib}
\usepackage{ifpdf}
\usepackage{placeins}
\usepackage{import} 
\usepackage{amsmath}
\usepackage{xcolor}
\usepackage{soul}
\usepackage{graphicx} 
\usepackage{makecell}
\usepackage{multirow}
\usepackage{graphicx}
\usepackage{float}
\usepackage{amsfonts}
\usepackage[T1]{fontenc}

\DeclareGraphicsExtensions{.pdf,.png,.jpg}

\begin{document}
\begin{frontmatter}

\title{Enhancing Biologically Inspired Hierarchical Temporal Memory with  
 Hardware-Accelerated Reflex Memory}

\author[1]{Pavia Bera\corref{cor1}}
\author[1]{Sabrina Hassan Moon}
\author[1]{Jennifer Adorno}
\author[1]{Dayane Alfenas Reis}
\author[1]{Sanjukta Bhanja}

\address[1]{University of South Florida}

\cortext[cor1]{Corresponding author: paviabera@usf.edu}

\maketitle


\begin{abstract}
 
 The rapid expansion of the Internet of Things (IoT) generates zettabytes of data that demand efficient unsupervised learning systems. Hierarchical Temporal Memory (HTM), a third-generation unsupervised AI algorithm, models the neocortex of the human brain by simulating columns of neurons to process and predict sequences. These neuron columns can memorize and infer sequences across multiple orders. While multiorder inferences offer robust predictive capabilities, they often come with significant computational overhead. The Sequence Memory (SM) component of HTM, which manages these inferences, encounters bottlenecks primarily due to its extensive programmable interconnects. In many cases, it has been observed that first-order temporal relationships have proven to be sufficient without any significant loss in efficiency. 
This paper introduces a Reflex Memory (RM) block, inspired by the Spinal Cord's working mechanisms, designed to accelerate the processing of first-order inferences. The RM block performs these inferences significantly faster than the SM. The integration of RM with HTM forms a system called the Accelerated Hierarchical Temporal Memory (AHTM), which processes repetitive information more efficiently than the original HTM while still supporting multiorder inferences. The experimental results demonstrate that the HTM predicts an event in 0.945 s, whereas the AHTM module does so in 0.125 s. Additionally, the hardware implementation of RM in a content-addressable memory (CAM) block, known as Hardware-Accelerated Hierarchical Temporal Memory (H-AHTM), predicts an event in just 0.094 s, significantly improving inference speed. Compared to the original algorithm \cite{bautista2020matlabhtm}, AHTM accelerates inference by up to 7.55x, while H-AHTM further enhances performance with a 10.10x speedup.

\end{abstract}
\end{frontmatter}

\section{Introduction}
\label{sec:intro}

The explosive growth in data generation from internet-connected devices has placed unprecedented demands on intelligent systems to process information efficiently and in real time. While traditional neural networks and other artificial intelligence (AI) algorithms have achieved remarkable success in domains like image recognition and natural language processing, they face significant limitations in adapting to continuous, dynamic streams of unlabeled data~\cite{laird2017standard}. These systems often rely on batch learning, requiring extensive training data, computational resources, and retraining when the input distribution changes. Moreover, their architectures are not inherently designed for real-time inference and adaptation, which are critical for applications like anomaly detection, robotics, and streaming data analysis.

In contrast,  Hierarchical Temporal Memory (HTM) offers a biologically inspired approach modeled on the structure and function of the human neocortex to address these challenges. HTM encodes and processes data by mimicking the brain's mechanisms for forming and recalling temporal patterns. HTM enables systems to adapt dynamically to new information, making it particularly suited for real-time applications ~\cite{hawkins2016neurons, ahmad2016neurons}. Its versatility is demonstrated through the Numenta Anomaly Benchmark (NAB), where HTM effectively processes several datasets such as Twitter volume, pay-per-click rates, network traffic, rogue agent detection, CPU utilization, temperature variations, and transportation data~\cite{lavin2015evaluating}. Furthermore, HTM's adaptability has been leveraged in fields such as medical image processing~\cite{zhou2018hierarchical}, wafer inspection~\cite{adam2018wafer}, biometric recognition~\cite{james2017htm}, robot localization~\cite{neubert2018sequence}, seismic activity prediction~\cite{micheletto2018using}, power load forecasting~\cite{osegi2018using}, and environmental monitoring~\cite{zyarah2020end}. These diverse applications underscore HTM's faithful emulation of the neocortex's mechanisms.

\begin{figure}[t]
\centering
  \includegraphics[width=1\columnwidth]{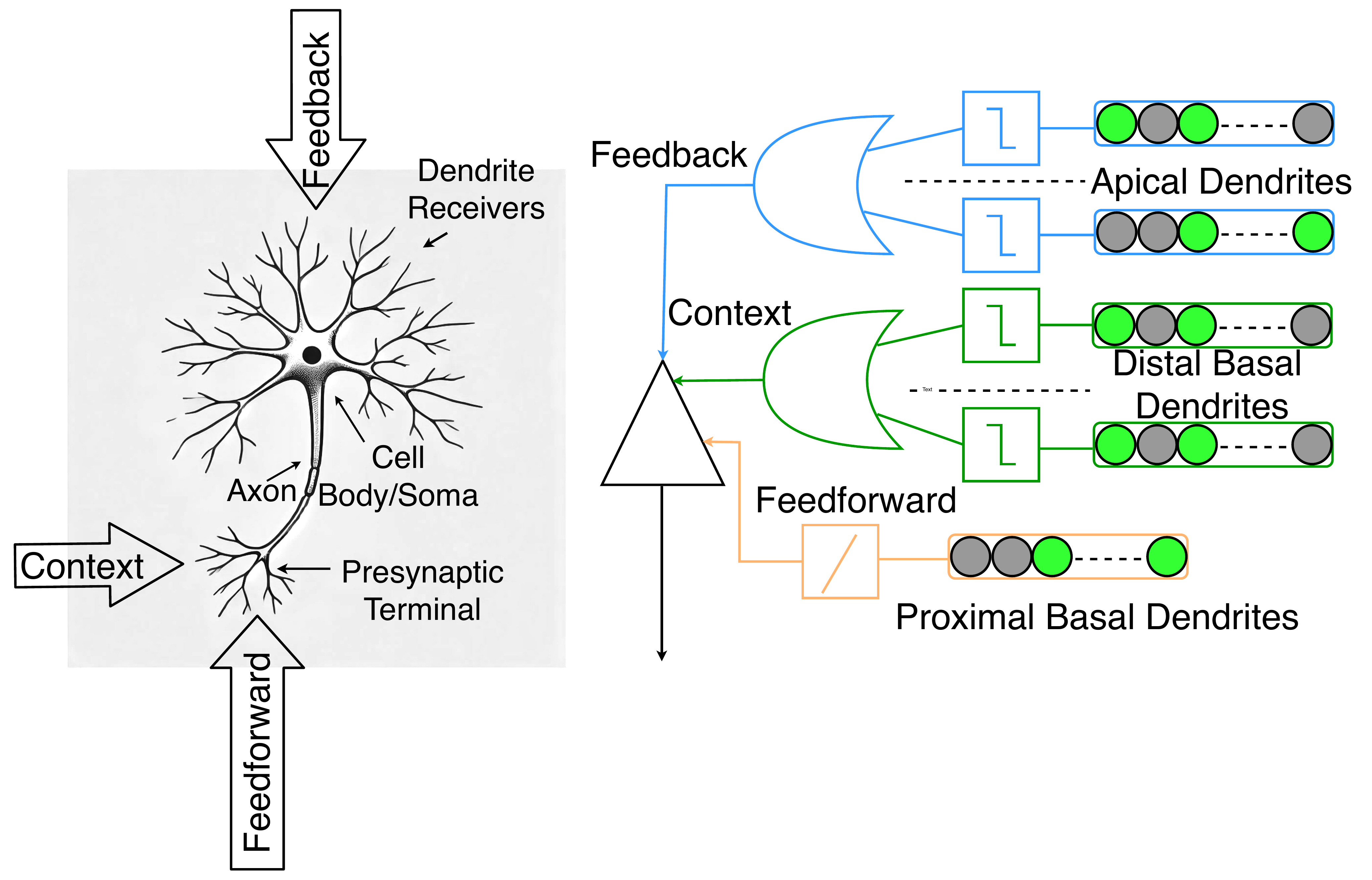}
\vspace{-4ex}
\caption{Comparison of biological and HTM neurons, highlighting feedback, feedforward, and contextual connections in both systems.}
\vspace{-3ex}
\label{fig:Biological_HTM}
\end{figure}

At the heart of HTM lies the Sequence Memory (SM), a biologically inspired component that replicates the processing capabilities of the neocortex through pyramidal neurons. These neurons interact via proximal, distal, and apical dendrites to encode neural activity and transmit data across hierarchical layers~\cite{zyarah2020end, bautista2020matlabhtm} as shown in Figure~\ref{fig:Biological_HTM}. These layers form the foundation of human reasoning by learning and predicting sequential events~\cite{hawkins2016neurons, bautista2020matlabhtm, harris2015neocortical}. The neocortex processes repeated events based on their co-occurrence frequency, integrating inputs from motor, visuomotor, and visuoperceptual sensors~\cite{laird2017standard, hawkins2016neurons, ahmad2016neurons}. Similarly, the SM establishes temporal connections between events by assigning sparse cells to memorize inputs~\cite{hawkins2016neurons, zyarah2020end}, enabling accurate online predictions without the need for batch processing~\cite{hawkins2016neurons, harris2015neocortical}. This allows HTM to excel in temporal associations, a critical capability for understanding and predicting dynamic data. However, the computational demands of SM, especially in hardware implementations, have traditionally posed a challenge. Maintaining dense interconnections and performing frequent updates makes achieving high-speed, resource-efficient processing challenging. Despite significant advancements to enhance HTM speed~\cite{hawkins2016neurons, zyarah2020end, bautista2020matlabhtm, krestinskaya2018hierarchical, cui2017htm}, its reactivity still falls short of centisecond processing demands~\cite{zhu2015segdeepm}.

\sloppy
To address these challenges, we introduce Reflex Memory (RM) as a lightweight, hardware-optimized component that complements the SM. Inspired by the brain’s ability to respond swiftly to familiar stimuli with minimal cognitive effort, RM is designed to process redundant, repetitive events efficiently. Biological reflex systems, such as those mediated by the spinal cord, enable ultra-fast responses by bypassing higher cognitive processing in the neocortex. Reflex arcs, which involve direct sensory-motor loops, allow the nervous system to react to familiar stimuli with minimal latency, optimizing for speed rather than complex inference. Similarly, in habitual learning, the basal ganglia reinforce frequently occurring motor patterns to reduce computational burden in decision-making \cite{graybiel1998basal,seger2011habit}. Unlike the SM, which handles complex, multi-order sequences, RM focuses on first-order inferences—predicting the most frequent events based on their co-occurrence history. This mirrors the biological principle of implicit memory formation, where repetitive stimuli lead to strengthened synaptic pathways in lower-order neural circuits, enabling rapid and automatic responses without invoking full cortical processing \cite{wikipedia2024procedural}. By specializing in this task, RM significantly reduces computational overhead for common patterns, offloading the SM and accelerating overall system performance. 

This optimization aligns with the broader understanding of memory in the brain. Multiple authors agree that intelligence relies on three primary memory systems: short-term, long-term, and working memory~\cite{doyon1998role,stevens2012role,camos2018not}. Generally, short-term memory processes information before passing it to the long-term system, which regularly contains 10,000 to 30,000 synapses per neuron~\cite{malenka1999long}. While Hierarchical Temporal Memory (HTM) retains valuable information in the temporal domain~\cite{zyarah2020end}, not all instances require the full computational power of Sequence Memory (SM). Our research introduces an enhanced approach, \textbf{Accelerated Hierarchical Temporal Memory (AHTM),} which builds on HTM by selectively offloading redundant data to Reflex Memory (RM) for faster response. RM acts as a biologically inspired fast-memory mechanism, similar to the role of basal ganglia in habit formation and the spinal cord in reflexive learning~\cite{doyon1998role,malenka1999long}. This enables real-time automation with minimal cognitive effort while maintaining the adaptive capabilities of HTM.

The main contributions presented in this paper are:

\begin{itemize}
    \item \textbf{AHTM:} Our novel AHTM architecture introduces Reflex Memory (RM)---a dynamically updated, fixed-sized dictionary-matrix algorithm designed to forecast the most frequent events without invoking the Sequence Memory (SM). RM is complemented by a Control Unit (CU) that mimics attention by selecting the appropriate prediction module and reinforcing learning between RM and SM. An online learning mechanism is also incorporated to adaptively update both RM and SM in real time, thereby enhancing system performance for dynamic and evolving data streams.
    \item \textbf{H-AHTM:} Building on AHTM, H-AHTM further integrates content-addressable memory (CAM) to accelerate RM and optimize event processing for sub-centisecond responsiveness.
\end{itemize}

Our evaluation of diverse financial datasets shows that both AHTM and H-AHTM achieve anomaly detection accuracy on par with traditional HTM models while significantly reducing inference latency. The results reveal that our approaches maintain robust precision, recall, F1-scores, and ROC-AUC values, with H-AHTM offering even greater speed improvements on high-frequency data. These findings highlight the potential of our methods as scalable, real-time solutions for financial anomaly detection and forecasting.

The rest of the paper is organized as follows: Section~\ref{sec:background} covers basic background on the HTM, including the Encoder, SP, SM, and CAM design. Section~\ref{sec:reflex} details the proposed architecture, highlighting RM and the CU. RM enhances SM efficiency by handling redundant inputs and integrates a CAM cell for sub-centisecond event processing. An online learning mechanism continuously updates RM and SM for adaptive learning. Section~\ref{sec:Method} outlines RM evaluation metrics, Section~\ref{sec:results} presents results, and Section~\ref{sec:Relatedwork} critically reviews relevant literature, identifying the limitations of existing approaches and clearly demonstrating the enhancements introduced by our method. Section~\ref{sec:conclu} concludes with future directions.

\section{Background}
\label{sec:background}

\begin{figure*}[t]
    \centering
    \includegraphics[width=1\textwidth,keepaspectratio]{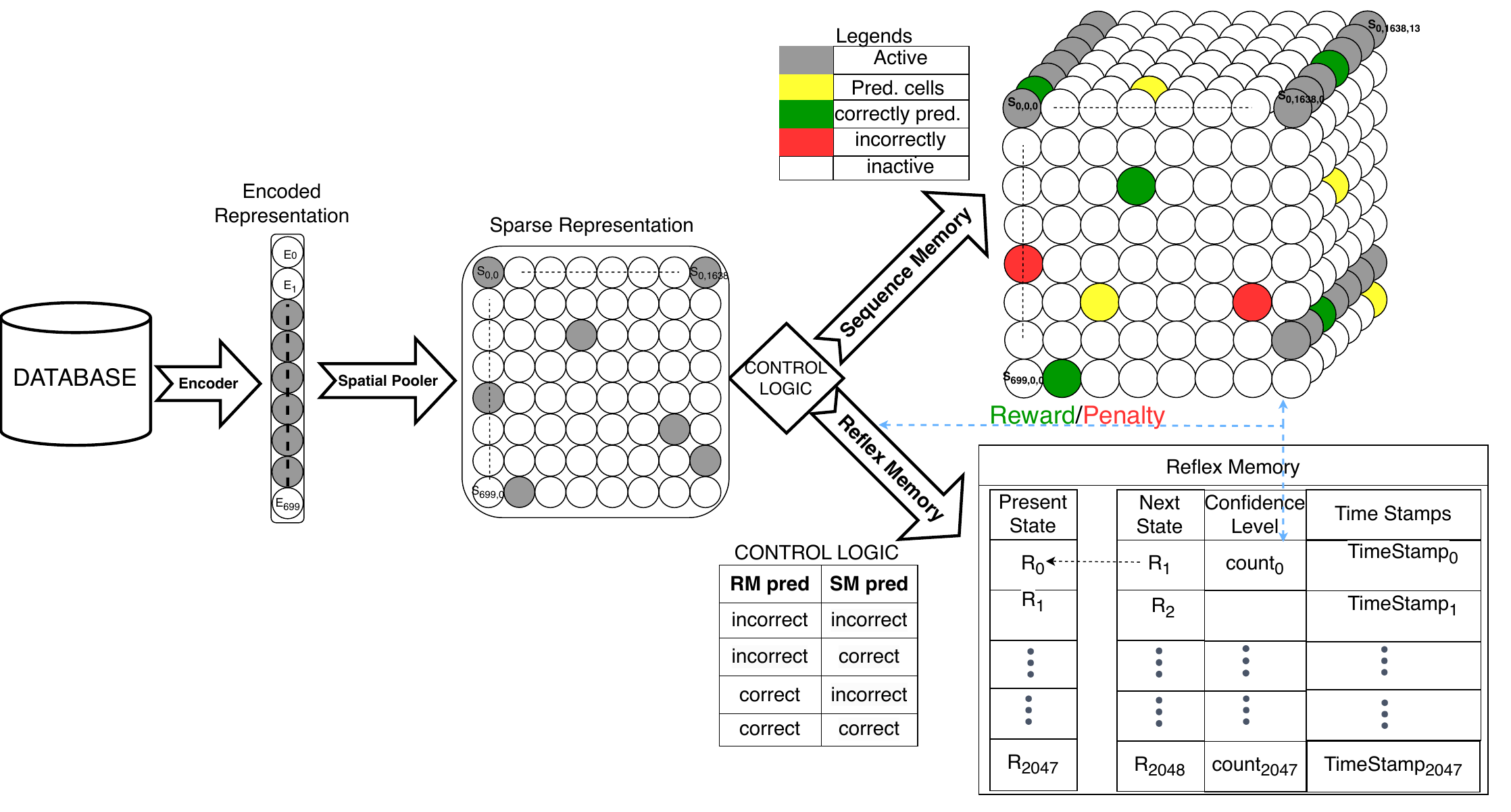}
    \vspace{-2ex}
    \caption{Generic Hierarchical Temporal Memory (HTM) Architecture. The architecture consists of a data flow pipeline starting from a database, which provides input to the Encoder. The Encoder translates raw input into a Symmetrically Encoded Representation (E), capturing cyclic features for processing by the Spatial Pooler (SP). The Control Unit coordinates operations, guiding the output to either the Sequence Memory (SM) for temporal learning or Reflex Memory (RM) for hardware-accelerated first-order computations. This structure enables time-efficient spatio-temporal learning and accurate prediction.}
    \label{fig:htm_architecture}
\end{figure*}

Hierarchical Temporal Memory (HTM) is a model inspired by the structure and function of the neocortex in the human brain. In HTM, memory is key to enabling systems to learn patterns and make predictions. The HTM architecture as referenced in Figure~\ref{fig:htm_architecture} learns and recognizes sequential data, similar to processes in the brain. 

In traditional HTM, the data flows sequentially from \textbf{Encoder} to \textbf{Spatial Pooler} (SP) to \textbf{Sequence Memory} (SM). Researchers continue to optimize the SP and SM for hardware implementations. The SP encodes information into a Sparse Distributed Representation (SDR) based on input from the encoder~\cite{purdy2016encoding}. The SM then learns temporal patterns from the encoded SDRs, acting as procedural memory and learning through long-term potentiation (LTP)~\cite{hawkins2016neurons,malenka1999long}. Together, these modules create a spatio-temporal representation of the data.
Our work, in turn, explores Compute-in-Memory (CiM) architectures based on \textbf{Content Addressable Memory} (CAM) due to its efficient memory access and pattern retrieval.

\subsection{Encoder}

In HTM, the encoder plays a crucial role in transforming raw input data into a SDR to ensure compatibility with the HTM framework. HTM systems rely on encoders to convert raw input data into SDRs, which serve as the primary format for further processing~\cite{ahmad2016neurons,cui2016continuous}. SDRs are binary arrays where each active bit encodes semantic meaning that facilitate efficient storage, pattern recognition, and noise robustness. The encoder's role is to produce an initial representation of the data that captures its essential features in a fixed-dimensional format, ready for processing by downstream components. Unlike the SP, which enforces a strict level of sparsity, the encoder primarily focuses on preserving semantic relationships in the data. While some encoders may not guarantee sparsity, most practical SDR encodings used in HTM are sparse to maintain their robustness and generalization properties ~\cite{purdy2016encoding}.

Eq. \ref{eq_1} describes the encoder mathematical function:

\begin{equation}  E: X \to \{0,1\}^{n}
\label{eq_1}
\end{equation}

Where \( X \) is the input space, \( \mathbf\{0,1\}^n \) is the \( n \)-dimensional binary SDR output space, and \( E(x) \) denotes the SDR representation for input \( x \in X \). For any two inputs \( x_1, x_2 \in X \), the encoder ensures that their similarity is preserved in the output space. This can be quantified by the Hamming similarity of the output SDRs (Eq. \ref{eq_2}):

\begin{equation}
\text{Sim}(E(x_1), E(x_2)) = \frac{1}{n} \sum_{i=1}^{n} \mathbb{I}[E(x_1)_i = E(x_2)_i],
\label{eq_2}
\end{equation}

Where \( \mathbb{I}[\cdot] \) is the indicator function. Higher similarity values indicate that the inputs \( x_1 \) and \( x_2 \) are semantically related.

\begin{figure*}[h]
    \centering
    \includegraphics[width=\columnwidth]{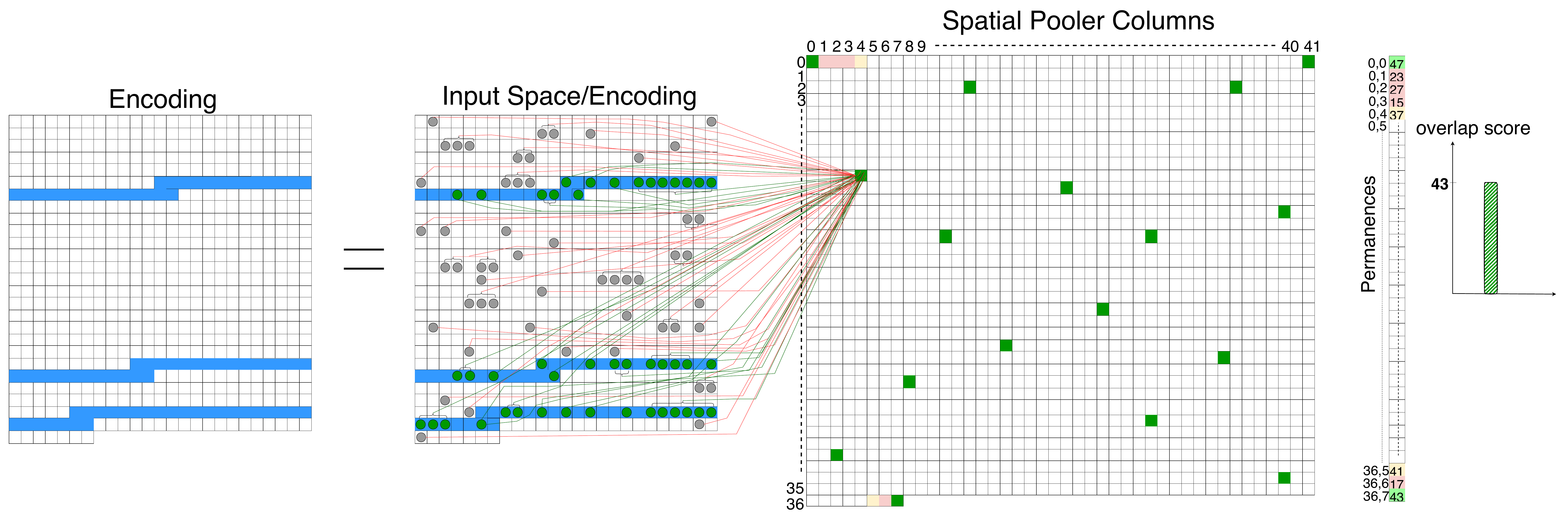}
    \vspace{-2ex}
    \caption{The SP transforms the encoded input vector \(E\) (blue bits) into a SDR \(S\), which is then processed by the SM. Each Input/Encoding space bit can potentially connect to any minicolumns in the Spatial Pooler. In the visualization, colors indicate different states: Green circles represent active bits in the input space that overlap with the encoded input, while grey circles denote inactive bits that fall outside the encoded space. Green synapses represent active connections within the encoded vectors, whereas red synapses indicate inactive connections in the current iteration. An accumulator $(\Sigma)$ counts the number of active synapses contributing to the overlap score. If the overlap score exceeds the threshold of 43, the corresponding minicolumn in \(SP\) is activated (ON).}

    \label{fig:sp}
\end{figure*}

The encoded SDRs serve as input to the SP, which refines these representations by enforcing sparsity, boosting underrepresented features, and enhancing noise robustness. The leftmost matrix in Figure~\ref{fig:sp} illustrates the initial encoded matrix generated by the encoder, preserving the key semantic features of the raw data. This matrix is then processed by the Spatial Pooler, which performs spatial encoding and transforms it into a sparse representation suitable for further temporal learning.

Encoders are adaptable and can be tailored to specific data types, such as numerical, categorical, or geospatial data. For example, a numerical encoder may map continuous ranges of values into overlapping or non-overlapping binary patterns, depending on the application. This adaptability ensures that HTM systems can process a wide variety of data inputs, bridging the gap between raw input and the sophisticated spatio-temporal processing performed by the SP and SM.

\subsection{Spatial Pooler}
The Spatial Pooler (SP) refines the initial SDR produced by the encoder \( E \), transforming it into a sparse and stable representation suitable for downstream temporal learning. While both the encoder and the SP output SDRs, their functions are fundamentally different. The encoder converts raw input data (e.g., numbers, categories, images) into an SDR that preserves semantic similarity, whereas the SP enforces sparsity and enhances robustness, ensuring that only the most relevant bits remain active. The output SDR from the SP exhibits the following key properties:
\begin{itemize}
    \item \textbf{Enforced sparsity}: Only a fixed percentage (typically \(\sim2-5\%\)) of bits are active, optimizing computational efficiency.
    \item \textbf{Noise robustness}: Similar input patterns produce SDRs with proportional overlaps, in their active (\(ON\)) bits~\cite{zyarah2020end,bautista2020matlabhtm,ahmad2016neurons} aiding generalization.
    \item \textbf{Stable representations}: The SP selects and maintains the most significant features in the input, preventing instability due to minor variations.
\end{itemize}
Each column in the output SDR \( S \) is activated based on a subset of bits in the input SDR \( E \), as depicted in Figure~\ref{fig:sp}. The connections between \( S \) and \( E \) are governed by a matrix \( SP \), where each column \( j \) maintains a set of potential synapses that connect it to multiple bits in \( E \), each with an associated permanence value. At initialization, the SP assigns random permanence values in the range \([0,1]\). These values define the strength of connections between the columns and input bits. Hebbian-like learning is then applied iteratively to adjust these permanence values, reinforcing active connections while weakening inactive ones~\cite{hawkins2016neurons,zyarah2020end}.

The computation of the output SDR \( S \) employs a k-Winners-Take-All (kWTA) selection mechanism, ensuring that only the most responsive columns become active. Mathematically, the SP computes overlap scores for each column \( j \) (Eq. \ref{sp_1}):
\begin{equation}
\label{sp_1}
O_j = \sum_{i} SP_{i,j} \cdot E_i
\end{equation}

Where \( SP_{i,j} \) represents the permanence matrix, and \( E_i \) is the encoded SDR input. The \textbf{kWTA function} then selects the top \( k \) most active columns:

\begin{equation}
S_j =
\begin{cases}
1, & \text{if } O_j \text{ is among the top } k \text{ largest overlaps} \\
0, & \text{otherwise}
\end{cases}
\end{equation}

This process ensures sparsity and feature selection by filtering out weakly active columns. To update synaptic permanence values, the following rule is applied (Eq. \ref{sp_2}):

\begin{equation}
\label{sp_2}
SP_{i,j} = SP_{i,j} + \alpha \cdot (2S_j - 1) \cdot E_i
\end{equation}

Where \( \alpha \) is the learning rate. This adjustment ensures that columns strengthen their connections to frequently co-active input bits while weakening unused connections. The final output SDR \( S \) is then passed to the SM, where temporal patterns are learned.


\begin{figure*}[t]
    \centering
    \includegraphics[width=\columnwidth]{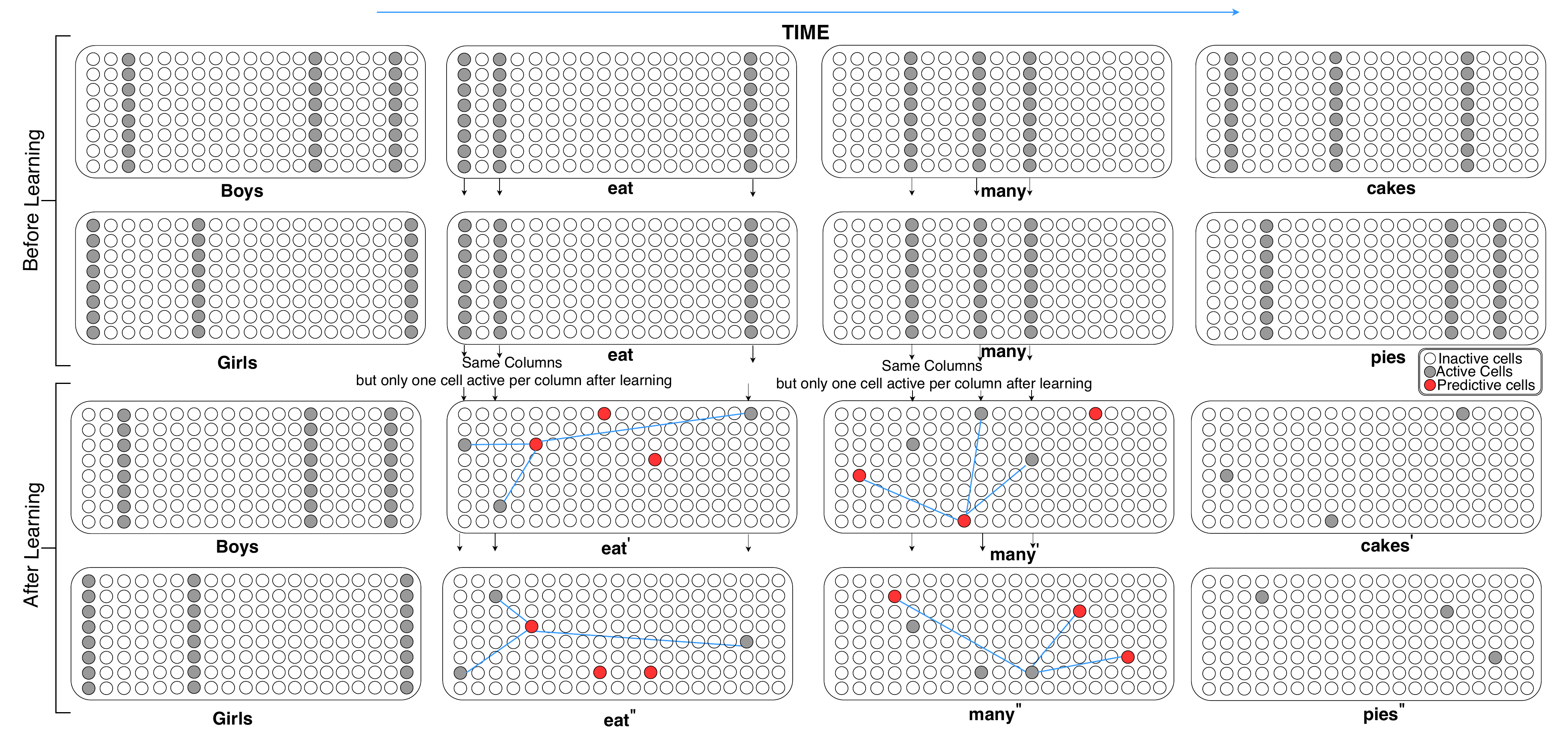}
    \caption{Illustration of SM operations. Each mini-column (MC) contains multiple cells, where only one cell becomes active after learning. The diagram shows the transition from the state before learning (with multiple active cells in a column) to the state after learning (with a single predictive cell per column). Predictive cells represent anticipated sequences, while active cells correspond to the current input. This mechanism enables the Sequence Memory to capture temporal patterns and make accurate predictions.}
    \label{fig:tm_diag}
\end{figure*}

\subsection{Sequence Memory}

The Sequence Memory (SM) acts as the procedural memory in the neocortex, establishing temporal connections between events through postsynaptic learning~\cite{hawkins2016neurons,doyon1998role,malenka1999long}. Mini-columns (MCs) in SM replicate the functionality of layer `$2/3$' of the neocortex~\cite{bautista2020matlabhtm}. Each MC contains \( p \) cells, all initially responding to the same spatial feature. Through learning, each cell specializes in recognizing different temporal contexts, allowing the system to predict sequences. This temporal encoding facilitates event forecasting but results in a large number of connections. SM neurons form temporal associations, strengthening connections via Hebbian-like learning. Predictive neurons reinforce dendritic connections to previously active cells, increasing the likelihood of accurate sequence recall.

The SM learns temporal patterns by associating sequential events with distinct cell activations. As illustrated in Figure~\ref{fig:tm_diag}, before learning, the system processes sequences such as "boys eat many cakes" or "girls eat many pies" by activating all cells in relevant MCs, as they have not yet specialized.
After learning, a single predictive cell per MC becomes active, representing a specific temporal context. Notably, common words like ``many" and ``eat" activate the same mini-columns across both sequences, as they share similar spatial features. This shared activation allows the system to generalize patterns while still differentiating temporal contexts.

After learning ``boys eat many cakes", the system refines the temporal pattern to ``boys $\text{eat}^{\prime}$ $\text{many}^{\prime}$ cakes," where the sequence now predicts cakes following $\text{many}^{\prime}$. Similarly, for ``girls eat many pies" the pattern transforms into ``girls $\text{eat}^{\prime\prime}$ $\text{many}^{\prime\prime}$ pies", associating pies with $\text{many}^{\prime\prime}$. Figure~\ref{fig:tm_diag} illustrates the process where multiple active cells in each MC before learning are reduced to a single predictive cell per column after learning. This mechanism enables the system to accurately anticipate temporal dependencies, even when sequences share overlapping spatial features, e.g., ``many" or ``eat".


The SM operates on a matrix \( SM \), where each cell \( SM_{j,k} \) at coordinates \((j,k)\) contains three key matrices: \( Dparents \), \( Dperms \), and \( ONparents \). Together, these matrices represent the dendrites within a cell. The system \emph{predicts} events by computing rows of these matrices:
\begin{itemize}
    \item \textbf{\( Dparents \)}: Stores the coordinates \((j,k)\) of \( ON \) cells that activated \( SM_{j,k} \) in previous iterations.
    \item \textbf{\( Dperms \)}: Contains the permanence values for the dendritic connections in \( Dparents \), with values in the range \([0,1]\). These values are initialized randomly.
    \item \textbf{\( ONparents \)}: A binary matrix that indicates the current state of connections in \( Dparents \), dictating which synapses can vote for activation.
\end{itemize}

When a row meets the threshold of \( ON \) synapses, it triggers an action potential in \( SM_{j,k} \). The vector \( Dwinner \) holds the highest-scoring dendrite of each cell, computed as:

\begin{equation}
    D_{\text{winner}} = (\text{ONparents} \circ \mathbb{I}(D_{\text{perms}} \geq \theta))
    \label{winner}
\end{equation}

Here, \( \theta \) is the voting threshold for synapses. The complexity of Eq.~\ref{winner} is \( O((d \cdot syn^2) \cdot r \cdot p) \), where \( d \) is the number of dendrites, \( syn \) is the number of synapses, \( r \) is the number of MCs, and \( p \) is the number of neurons per MC. Each MC selects its most confident cell through a winner-take-all (kWTA) operation~\cite{hawkins2016neurons}. The resulting array of top scores forms the predicting vector \( Rpredicted \), which is compared to the actual input vector \( R^i \). A significant difference between \( Rpredicted \) and \( R^i \) signals an anomaly.

The system learns by updating two key matrices:
\begin{itemize}
    \item \textbf{\( SM_{\text{predict}} \)}: Identifies cells that predicted the next activation, marking them with \( 1 \)-values.
    \item \textbf{\( SM_{\text{learn}} \)}: Determines learning cells by applying the Hadamard product (\( \circ \)) between \( R \) and each row of \( SM_{\text{predict}} \), computed as:
    \begin{equation}
        SM_{\text{learn}} = kWTA \left( \begin{bmatrix}
            R \circ SM_{\text{predict},1,*} \\
            R \circ SM_{\text{predict},2,*} \\
            \vdots \\
            R \circ SM_{\text{predict},o,*}
        \end{bmatrix} \right)
    \end{equation}
\end{itemize}

This computation has a complexity of \( O(r^2) \). Winning MCs reinforce synaptic permanences in \( Dperms \), following long-term potentiation (LTP), while incorrect MCs reduce permanences, mimicking long-term depression. This process ensures recognition of frequent patterns while deprioritizing rare ones. The learning cells excite the respective \( Dparents \), and the system iteratively predicts the next event.

\subsection{Content Addressable Memory (CAM)}
CAM has been widely employed for addressing the processor-memory bottleneck, notably enhancing scalability, speed, and power efficiency within CiM architectures \cite{narla22, moon2024afecam, yang2019ternary, yin2019_tcam}. To further support data-intensive applications, a novel design featuring an ultra-dense, energy-efficient single Ferroelectric Field Effect Transistor (FeFET) based CAM cell, AFeCAM was introduced in~\cite{moon2024afecam}. 

\begin{figure}[H]
    \centering
    \includegraphics[width=\columnwidth]{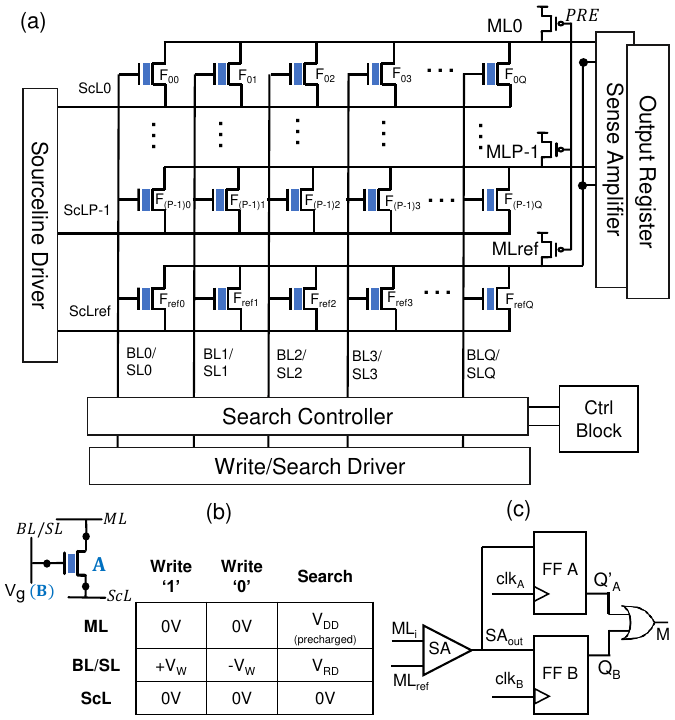}
    \vspace{-2ex}
    \caption{Architecture of an (a) AFeCAM subarray (b) the AFeCAM cell design, and an (c) output register (Adapted from \cite{moon2024afecam}).}
    \label{fig:subarray}
\end{figure}

This design optimizes storage and search capabilities, as depicted in Figure~\ref{fig:subarray}(a), by organizing P-1 words across P-1 rows with an additional reference row to facilitate search operation. Each row contains Q cells storing a single bit. The architecture has horizontal matchlines ($ML$) and sourcelines ($ScL$), and vertical bitline/searchline ($BL/SL$) connections within a subarray.

The cell-level view, write and search schemes, and periphery circuits for this CAM design are depicted in Figure~\ref{fig:subarray}(b). To write in an AFeCAM subarray, at first, the sourceline driver connects the selected $ScL_{i}$ to 0 V. Then the write driver applies a voltage of $+V_{w}$ ($-V_{w}$) to the $BL_{j}$ to write a logic `1' (`0') in the $j^{th}$ cell of row $i$, i.e., in cell $F_{ij}$. The output register, illustrated in Figure~\ref{fig:subarray}(c), is equipped with a comparator based voltage sense amplifier (SA), two flip-flops (FF A and FF B) and a two-input OR gate. The search operation consists of two phases: pre-search and search phase. At the beginning of the search operation $ML$s are precharged to $V_{DD}$. In the pre-search phase, a mismatch is detected when the cells store `1', but `0' is being searched. Conversely, during the search phase, a mismatch occurs when the cells store `0' while `1' is being searched. Finally, the output node M identifies the result of an exact match.


\section{Accelerated HTM: AHTM and H-AHTM Innovations}
\label{sec:reflex}

In this section, we present our accelerated HTM framework, which comprises two complementary approaches: AHTM and H-AHTM. AHTM introduces a novel mechanism—Reflex Memory (RM)—that leverages a lightweight, dictionary-matrix algorithm to rapidly predict frequent events without invoking the traditional Sequence Memory (SM). Building on this concept, H-AHTM further accelerates RM by integrating content-addressable memory (CAM) to achieve sub-centisecond responsiveness.

While RM is inspired by the rapid, automatic responses observed in biological systems, and CAM is well-known for its efficient search capabilities, our focus here is on how these components collectively enable real-time online learning and prediction. The following subsections detail the algorithm overview of RM, the control unit for selective attention, and the online learning mechanism, followed by an exploration of the CAM design and hardware mapping in H-AHTM.

\begin{figure}[t]
    \centering
    \includegraphics[width=1.0\columnwidth]{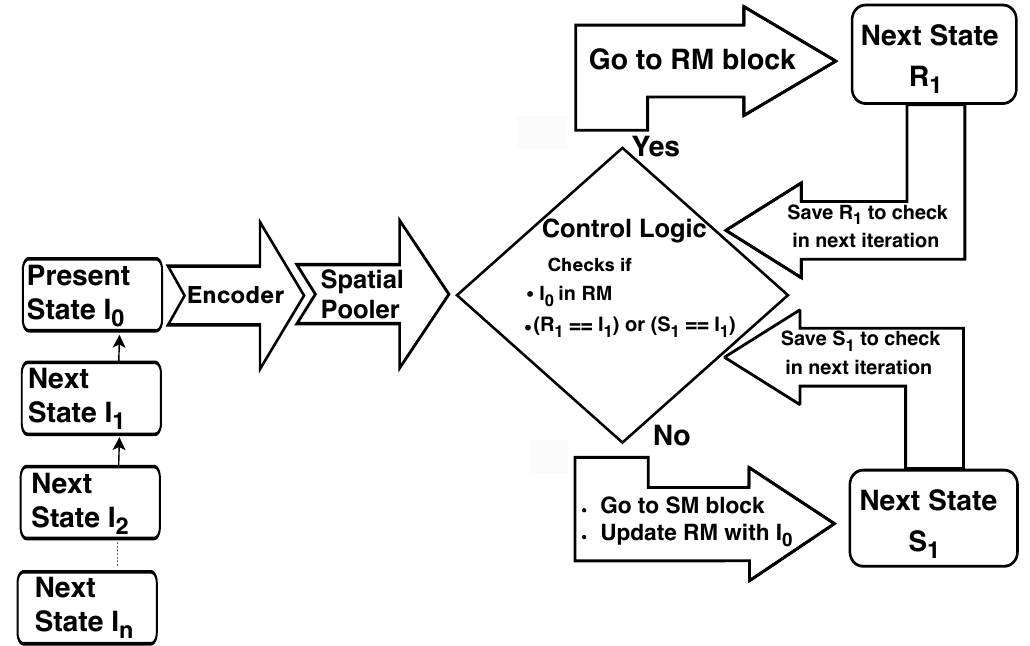}
    \vspace{-2ex}
    \caption{Interaction between RM, SM, and the SP, detailing the control logic that governs state transitions, input encoding, prediction generation, and experience consolidation.}
    \vspace{-2ex}
    \label{fig:online}
\end{figure}

\subsection{AHTM: Acceleration via Reflex Memory}


\textbf{RM Algorithm Overview:}
\label{subsec:algo_reflex}
The concept of Reflex Memory (RM) has been explored in neurological and cognitive science studies, where it is associated with automatic responses, maladaptive memory formation, and implicit learning~\cite{oyigeya2021reflex, pavlov1927conditioned, beritashvili1974image}. For instance, Oyigeya~\cite{oyigeya2021reflex} described RM as an implicit form of memory that encodes interoceptive experiences to influence motor and sensory responses, while classical studies by Pavlov~\cite{pavlov1927conditioned} and Beritashvili~\cite{beritashvili1974image} investigated how reflexive behaviors can be conditioned and guided by mental representations. However, prior work has not successfully implemented RM using HTM.

\par 
Our proposed approach addresses this gap by introducing a hardware-efficient RM mechanism that mimics the rapid, automatic responses seen in biological reflex systems, thereby enabling real-time learning and prediction. Inspired by the spinal knee-jerk reflex—a swift, involuntary response triggered by a tap on the patellar tendon—RM processes redundant data and predicts frequently occurring sequences with high speed and efficiency. As illustrated in Figure~\ref{fig:online}, this design bypasses the slower, more complex decision-making systems typical of traditional Sequence Memory (SM) operations~\cite{doyon1998role, stevens2012role, camos2018not}.

\par \vspace{\baselineskip}

\begin{figure}[t]
\centering
\includegraphics[width=\columnwidth,keepaspectratio]{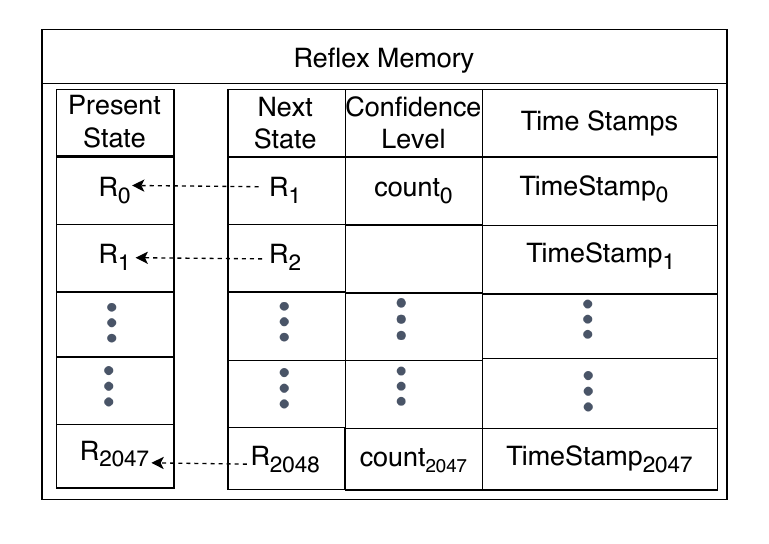} 
\vspace{-8ex}
\caption{The architecture and key components of RM.}
\label{overview}
\end{figure}

RM learns and predicts by associating frequently sequential events. Let \textit{present state} (\(R_1\)) and \textit{next state} (\(R_2\)) represent two consecutive occurrences. RM uses a dictionary structure to store these sequences. The RM module, illustrated in Figure~\ref{overview}, includes a dictionary \(D\) that maps each key (e.g., \(R_1\)) to a list of possible values (e.g., \(R_2, R_3\)) with associated recurrence counts, such as:

\begin{equation}
D: R_1 \to [R_2: 20, R_3: 50].
\end{equation}

RM predicts the value with the highest count (here, \(R_3\)) until a new value surpasses the count. For each prediction, RM updates the recurrence count of the observed value. If \(R_2\)'s count surpasses \(R_3\), \(D\) updates \(R_1\) to point to \(R_2\). 

To manage memory efficiently, RM has a fixed size of 2048 entries. This limit was chosen as a power of two to optimize memory allocation and addressing—balances capacity and efficiency, preventing excessive memory usage in online learning, where storing all data, including outdated patterns, would overwhelm the system. When full, RM evaluates entries based on last access time and recurrence counts, removing rarely used and low-frequency entries. This ensures RM remains adaptive and focused on processing the most relevant patterns for real-time predictions while also enabling the system to continuously learn new sequences without catastrophic forgetting -- a key requirement for online learning.
Additionally, the fixed size can be adjusted based on available resources.

\textbf{Control Unit for Selective Attention:}
\label{sec:control_unit}
With RM efficiently managing memory, the next challenge is deciding when to rely on RM versus SM for optimal predictions. To address this, we introduce the \textbf{Control Unit (CU)} for Selective Attention, which dynamically selects the best prediction module based on real-time performance while reinforcing learning.
The CU continuously monitors RM and SM’s performance, adapting based on prediction confidence and accuracy. This design mirrors the brain’s attentional processes, where rapid reflexive actions (subcortical processing) are balanced with complex decision-making (cortical learning).

To determine which module should be used for inference, the CU compares the summation of the last four anomaly scores from RM against the corresponding summation from SM. By default, RM is preferred due to its speed advantage, but if its accumulated anomaly score is higher than SM’s, the CU shifts reliance to SM for better accuracy. The decision to use the summation of the last four anomaly scores as a thresholding mechanism is based on the need for a balance between short-term adaptability and long-term stability. A shorter window (e.g., one or two scores) may lead to erratic switching between RM and SM, while a longer window (e.g., ten or more) could result in slow adaptation to changing patterns. Both RM and SM receive the same input in parallel, ensuring that inference remains synchronized. This mechanism allows RM to be leveraged for fast predictions while maintaining the robustness of SM for more challenging cases.

During training, the CU follows these memory update rules:

\begin{itemize}
    \item \textbf{RM Incorrect, SM Incorrect:}  
    \begin{itemize}
        \item Both RM and SM update their predictions.
        \item RM’s recurrence counts are adjusted to improve future predictions.
    \end{itemize}
    
    \item \textbf{RM Incorrect, SM Correct:}  
    \begin{itemize}
        \item SM’s output is used, bypassing RM.
        \item RM reduces its recurrence count for the incorrect prediction.
        \item RM is retrained to align with SM’s accurate predictions.
    \end{itemize}
    
    \item \textbf{RM Correct, SM Incorrect:}  
    \begin{itemize}
        \item RM’s prediction is used, bypassing SM.
        \item SM is trained with a regular confidence update to improve accuracy.
    \end{itemize}

    \item \textbf{RM Correct, SM Correct:}  
    \begin{itemize}
        \item RM’s output is used to expedite response time.
        \item SM is reinforced with a higher confidence update.
    \end{itemize}
\end{itemize}

By continuously adapting to changing conditions, the CU ensures an optimal balance between fast inference from RM and long-term accuracy from SM, enabling a robust and adaptive prediction system.

\textbf{Accelerated HTM for Online Learning: }
In today’s dynamic environment, systems must adapt and learn in real time—a challenge for traditional paradigms reliant on static datasets and offline training. Although the HTM framework is celebrated for its biologically inspired spatio-temporal learning, its reliance on static representations and the computationally intensive Sequence Memory (SM) limits its capacity for online learning.

Our framework addresses these challenges by integrating Reflex Memory (RM) for rapid predictions. This dual-memory design ensures robust Online learning by leveraging the spatio-temporal processing capabilities of the neocortex, facilitating rapid predictions and incremental learning of complex temporal sequences.

\begin{figure*}[t]
    \centering
    \includegraphics[width=\textwidth]{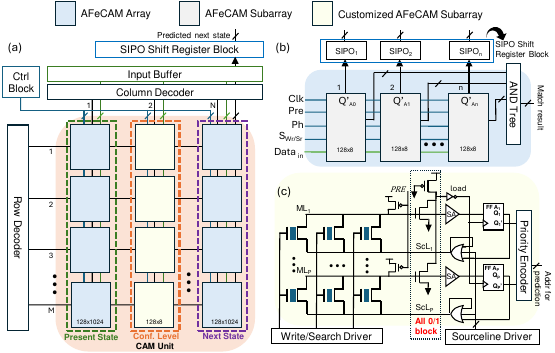}
    \caption{(a) An overview of Hardware Accelerated Hierarchical Temporal Memory (H-AHTM) architecture. (b) AFeCAM array for storing SDR in the present and next state block. (c) customized AFeCAM subarray design for confidence level block.}
    \label{fig:AHTM}
\end{figure*}


\subsection{H-AHTM: Acceleration via CAM-Enhanced Reflex Memory}
\label{subsec:AHAHTM}
Here, we describe our proposed hardware accelerated hierarchical temporal memory (H-AHTM) for RM (Sec.~\ref{subsec:algo_reflex}) based on CAM. Our design adopts the 1-FeFET CAM design from~\cite{moon2024afecam} (herein referred to as AFeCAM). The H-AHTM architecture,illustrated in Figure~\ref{fig:AHTM}(a), includes a CAM unit, decoders, registers, a buffer, and a control block to execute the functionalities of RM. Though~\cite{moon2024afecam} emphasizes using CAM for search operation only, we incorporate a new element, i.e., serial-in, parallel-out (SIPO) shift register, which enables the array to work as both CAM and memory unit. 

Below, we describe each of the components of our proposed H-AHTM architecture in detail.

    \textbf{CAM Unit:} In our design, we split the overall CAM architecture into three array stages as marked in Figure~\ref{fig:AHTM}(a) to store: a) Present State, b) Confidence Level and c) Next State depicted in green, orange and purple dashed rectangle respectively. The present and next state stages consist of $M$ AFeCAM arrays. Figure~\ref{fig:AHTM}(b) illustrates the AFeCAM array design, which contains $n$ AFeCAM subarrays sized $P \times Q$ and an AND tree. The AFeCAM subarrays are detailed in Sec. \ref{sec:background}, and in \cite{moon2024afecam}. To obtain the final search results from $n$ subarrays, all the match results are passed through an AND tree at the array level. The AND tree operates in a hierarchical manner, where the outputs from multiple subarrays are progressively reduced to a single output. Additionally, in the next state block, each subarray in the design is equipped with its dedicated register, resulting in a total of $n$ registers at the array level. 
    
    To support the prediction functionality of RM in this work as shown in Figure~\ref{fig:AHTM}(b), we propose to utilize the inverted output Q' of Flipflop-A (FFA) of the output register (Figure~\ref{fig:subarray}(c)) from each subarray. In each iteration, these outputs are directed to the register block, and predictions are made through the output node of this register block. This approach enables direct retrieval of stored states from the H-AHTM, facilitating efficient and reliable access to the prediction result. 
    
    The confidence level stage consists of $M \times1$ customized AFeCAM subarrays. The customized AFeCAM subarray is illustrated in Figure~\ref{fig:AHTM}(c). 
    In our design, we have incorporated a distributed pseudo-NMOS NOR gate, named the `all 0/1 block', which efficiently detects misses during search operations. This gate features a pull-up network (PMOS) that is constantly active, ensuring a default high state. The pull-down network, on the other hand, is connected to each $ML_i$. When a miss is detected in $ML$, it remains charged in pre-search phase, it biases the NMOS transistor in the pull-down network. This results in the output becoming 0, which indicates a miss. This mechanism is crucial for the row selection process during min/max operations (see Sec.~\ref{subsec:AHAHTM}(4)). The priority encoder gives the final valid address to identify the most frequently seen SDR and the row decoder activates the corresponding row in the next state stage, allowing the system to read out the stored SDR.

    \textbf{Decoders:} To access data across the H-AHTM architecture, we utilize both a row decoder and a column decoder for efficient memory addressing and retrieval. The row decoder is responsible for selecting and activating a specific AFeCAM array within the CAM unit to facilitate different RM operations. It takes a $\log_2(M)$-bit address as input and determines the row to be accessed for writing new data or predicting from stored SDRs. The column decoder takes a $\log_2(N)$-bit address to select which stage is activated within the architecture. 

    \textbf{Registers:} Our H-AHTM architecture integrates a SIPO shift register block containing $n$ $Q$-bit SIPO shift registers for $n$ AFeCAM subarray in the next state stage, enabling efficient serial-to-parallel data conversion. When the row decoder activates the desired prediction entry, an iterative predict operation is performed and the corresponding shift register receives data from the selected row. The AFeCAM subarray stores $Q$-bit data per row, which is loaded into the shift register bit by bit over consecutive clock cycles. After $Q$ clock cycles, the shift register completes loading the full $n$ $Q$-bit word, making it available for parallel readout from the register block.

    \textbf{Input Buffer and Control Block:} An input buffer regulates data writing operations across the array. A dedicated control block generates the clock, precharge (Pre), phase (Ph), and $S_{Wr/Sr}$ signals to facilitate the coordinated operation of each AFeCAM subarray within the array. This also checks match results from different stages to efficiently manage and execute each operation within the system.

The H-AHTM architecture supports five operations:

    \textbf{(1) Write:} To write the present or next state, as well as confidence level in H-AHTM, we follow the same write operation as described in Sec. \ref{sec:background}-D and in~\cite{moon2024afecam}. The column decoder activates corresponding stages in CAM unit while processing each stage.
    
    \textbf{(2) Search:} To search in the present state and next state, H-AHTM incorporates the two-phase search mechanism detailed in~\cite{moon2024afecam}. The column decoder selects the stage to perform the search operation. When processing, only one stage is activated. However, to update an instance with SDR $R_i$ and $R_{i+1}$, the column decoder simultaneously activates both the present and next state stages for searching.

    \textbf{(3) Update:} H-AHTM dynamically updates low-frequency entries with the most recent data based on the last access time and recurrence count. This process involves writing `0' to all cells in the target row ($i_{th}$ row) that require updating. AHTM evaluates the time stamps and sends the less frequently seen SDRs i.e., $R_i$ and $R_{i+1}$ for one instance as query to the CAM unit. $R_i$ and $R_{i+1}$ are searched in present and next state stages, respectively. Both are overwritten with `0' and the corresponding confidence level is reset to `0'. This operation resets the cells to their initial state, making them ready to be overwritten with the new instance.
\begin{figure}[b]
    \centering
    \includegraphics[width=1\columnwidth, height=100pt]{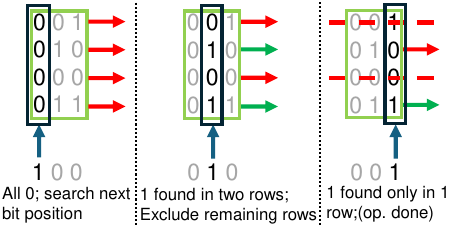}
    \caption{An example for finding maximum within a CAM block.}
    \label{fig:min/max}
\end{figure}

    \textbf{(4) Min/Max:} Min/max operations are essential for evaluating confidence levels within the dataset. They utilize an iterative bitwise search algorithm to determine the maximum (or minimum) value. Figure~\ref{fig:min/max} illustrates a max operation within the confidence level stage. The algorithm begins by searching for `1' in the most significant bit (MSB) position across all the rows. If a given bit position contains `0' in all levels, the second iteration proceeds to the next lower bit position. In cases where certain rows contain `1's while others store `0's, the rows storing `0's are excluded from subsequent iterations, referred to as row exclusion. 
    On the other hand, rows containing `1's will be precharged for the next iteration of the search operation. This iterative process continues until a single row remains to store `1', signaling the termination of the search operation. To find the minimum confidence level, we can search for `0's using the same process.

    \label{subsubsec:minmax}

    \textbf{(5) Prediction:} To predict the next state, we propose a prediction operation in the CAM, leveraging the capabilities of AFeCAM. Utilizing the presearch phase as described in~\cite{moon2024afecam}, we can execute this operation. The inverted output Q' of FF A is connected to the input of SIPO shift register. The parallel output from n registers allows us to predict an SDR $R_{i+1}$ from the next state stage. Since our AFeCAM subarray stores Q bit per row, the prediction process requires Q precharge-presearch cycles to predict the next state within the CAM block. 
    
\subsection{Hardware Mapping of RM}

Here, we describe the mapping of the RM algorithm described in Sec. \ref{subsec:algo_reflex} to H-AHTM. Our architecture operates through well-defined instructions that enable interaction between the present state, next state, and confidence level stages, ensuring optimized data storage and retrieval. In the following, we detail the mapped operations involved in this process.

\textbf{Present State Processing:} 
At first, an SDR $R_i$ is loaded into the present state stage of our H-AHTM. When $R_i$ is generated by the SP, the column decoder activates the present state stage, and a search operation is initiated to check whether the SDR already exists in the stage. If no match is found, this indicates that the SDR is a new instance. In this case, the controller activates the write mode, and row decoder will activate an array to store the SDR. This ensures that previously unseen patterns are recorded for future reference. However, if a match is found, indicating that the $R_i$ was previously seen, the system utilizes the match result from the AND tree to access the corresponding confidence level stage.

\textbf{Confidence Level Processing:} This stage plays a crucial role in determining the most frequent SDR for accurate prediction. The processing begins by analyzing the match results from the present state stage. If multiple matches are found for $R_i$, the system must determine which stored SDR in the next state stage has the highest frequency of occurrence. This is achieved through an iterative min/max operation as described in Sec.~\ref{subsec:AHAHTM}(4), where the match results are evaluated progressively to identify the SDR with the maximum frequency.  This hierarchical approach ensures that the most reliable prediction is selected based on previous occurrences. However, if a single match is found in the present state stage, the control block allows the system to bypass confidence level processing entirely. In this case, the identified row in the next state stage is directly activated, and the corresponding SDR $R_{i+1}$ is read out from the shift register block.

\textbf{Next State Processing:} Once the maximum frequent confidence level is determined, the system activates the corresponding row in the next state stage to retrieve the stored value and then sends it to the control unit of the AHTM. However, when a present state $R_i$ is encountered for the first time, there is no corresponding next state $R_{i+1}$ stored in the block. In this case, the system waits for the SP to generate $R_{i+1}$ as part of the learning process. Once the SP provides $R_{i+1}$, the controller activates the write mode and stores $R_{i+1}$ in the same row address as $R_i$ within the next state stage. As the system continuously learns and stores new SDRs, the memory eventually reaches its storage capacity. The software periodically scans the stored entries to detect the least frequently occurring $R_{i+1}$, identifying the entry that needs to be replaced. Once the lowest frequent SDR is identified, an update operation is performed as described in Sec.~\ref{subsec:AHAHTM}(3). 
This ensures efficient memory utilization without disrupting frequently accessed $R_{i+1}$.

\begin{table}[t]
\centering
\caption{Dataset Summary}
\resizebox{\columnwidth}{!}{%
\begin{tabular}{|c|c|c|c|c|}
\hline
\begin{tabular}[c]{@{}c@{}}\textbf{Dataset}\\ \textbf{Name}\end{tabular}        & \begin{tabular}[c]{@{}c@{}}\textbf{Temporal}\\ \textbf{Resolution}\end{tabular} & \begin{tabular}[c]{@{}c@{}}\textbf{Data}\\ \textbf{Points}\end{tabular} & \textbf{Features}          & \begin{tabular}[c]{@{}c@{}}\textbf{Time}\\ \textbf{Span}\end{tabular}           \\ \hline

Dow Jones             & Weekly                      & 2082                & Value     & 9/2/1977 to 8/29/2017       \\ \hline
NASDAQ                & Weekly                      & 2083                & Value    & 9/2/1977 to 8/29/2017        \\ \hline
 S\&P 500              & Weekly                      & 2084                & Value     & 9/2/1977 to 8/29/2017        \\ \hline
Gold Prices          & Monthly                       & 842                & Price                     & 01/1950 to 01/2020        \\ \hline
S\&P 500             & Monthly                     & 1766                & Several$\mathbf{{}^\star}$     & 1/1/1871 to 4/1/2018        \\ \hline
VIX Close            & Monthly                     & 4053                & VIX Close                     & 1/2/2004 to 2/7/2020       \\ \hline
VIX High             & Monthly                     & 4053                & VIX High                      & 1/2/2004 to 2/7/2020        \\ \hline
VIX Low              & Monthly                     & 4053                & VIX Low                       & 1/2/2004 to 2/7/2020        \\ \hline
VIX Open             & Monthly                     & 4053                & VIX Open                      & 1/2/2004 to 2/7/2020        \\ \hline

VIX Close             & Daily                       & 4053                & Price                     &  1/2/2004 to 2/7/2020           \\ \hline
VIX High             & Daily                       & 4053                & Price                     &  1/2/2004 to 2/7/2020         \\ \hline
VIX Low             & Daily                       & 4053                & Price                     &  1/2/2004 to 2/7/2020         \\ \hline
VIX Open             & Daily                       & 4053                & Price                     &  1/2/2004 to 2/7/2020         \\ \hline
Natural Gas            & Daily                       & 5802                & Price                     & 1/7/1997 to 2/3/2020        \\ \hline
Oil Prices             & Daily                       & 8303                & Price                     & 5/20/1987 to 2/3/2020        \\ \hline
\end{tabular}%
}
\label{dataset_summary}

\begin{flushright}
{\bf $\mathbf{{}^\star}$}Value, Dividend, Earnings, Consumer Price Index, \\ Long Interest Rate, Real Price, Real Dividend, Real Earnings. \\
\end{flushright}

\end{table}

\begin{table}[t]
\centering
\caption{Latency and Energy for Various Operations}

\begin{tabular}{|c|c|c|}
\hline
\textbf{Operation} & \textbf{Latency (ns)} & \textbf{Energy (fJ/bit)} \\ \hline
Write & 20 & 0.16 \\ \hline
Search & 0.25 & 0.22 \\ \hline
Update & 20.25 & 0.54 \\ \hline
Min/Max & 1.2 & 1.76 \\ \hline
Predict & 2.3 & 1.76 \\ \hline
\end{tabular}%
\label{tab:latency_energy}
\end{table}

\section{Evaluation}
\label{sec:Method}
In this section, we evaluate the performance of AHTM and H-AHTM based on two key metrics: anomaly detection and processing speed. Additionally, we discuss the accuracy vs. speed trade-off as a crucial consideration.


%

\subsection{Evaluation Setup}
To evaluate the performance of the proposed AHTM framework, a series of experiments were performed using financial datasets comprising various financial indicators (e.g., stock indices, commodity prices, and volatility indices) with temporal resolutions ranging from daily to monthly. Table~\ref{dataset_summary} summarizes the key attributes of each dataset. The experiments were conducted on a virtualized Linux environment equipped with an 8-core Intel i9 processor (3.0 GHz base clock, 5.8 GHz max clock), 10 GB of RAM, and Hyper-Threading enabled. Each dataset was processed ten times to compute average performance metrics.

The AHTM framework incorporates a software-only Reflex Memory (RM) that was benchmarked against HTM's Sequence Memory (SM). In contrast, the hardware-accelerated version, H-AHTM, integrates the RM via CAM and was benchmarked against both HTM's SM and AHTM's RM to assess comparative performance.

In addition to algorithmic evaluation, the hardware capabilities of H-AHTM were assessed by measuring energy consumption \& latency. At the circuit level, we utilize SPICE simulations to measure the energy and latency of the AFeCAM subarray within our H-AHTM architecture. These simulations leverage a FeFET multidomain model calibrated with experimental data from Intel’s 14 nm FinFET technology \cite{gupta2020temperature}. We implement the SIPO register, AND tree, and output register blocks using Verilog and evaluate their energy and latency through the synthesis in Cadence Genus RTL Compiler, using the NanGate 45-nm open-cell library \cite{knudsen2008nangate}.

To accommodate several datasets, we carefully design our H-AHTM architecture to support 2048 entries of 1024-bit SDR generated from the SP module. Consequently, the parameters used in our design are $n$=128, $M \times N$ = $16\times3$ and $P \times Q$ = $128 \times 8$. However, for confidence-level, a $M$ AFeCAM subarray is sufficient to store the resulting frequency. The energy and latency measurements for various H-AHTM-supported operations are summarized in Table~\ref{tab:latency_energy}.

Scalability was a key consideration in the design of AHTM and H-AHTM, ensuring efficient handling of increasing data volumes through Reflex Memory and CAM acceleration. The architecture ensures adaptability to larger datasets and real-time streaming scenarios by leveraging Reflex Memory and CAM acceleration.

\subsection{Evaluation Metrics}
The performance of AHTM and H-AHTM was evaluated based on the following key metrics:
\begin{itemize}
    \item \textbf{Anomaly Detection Accuracy:} This metric is quantified using Precision, Recall, F1-score, and ROC-AUC. Anomaly detection accuracy was evaluated using the anomaly raw score (ARS) defined as:
    \begin{equation}
    ARS^i = 1 - \frac{\text{nz}(R_{\text{predicted}} \cap R^i)}{\text{nz}(R^i)}
    \end{equation}
   Where $\text{nz}$ accounts for the non-zero elements in a vector~\cite{lavin2015evaluating}, $R_{\text{predicted}}$ is the predicted output from AHTM and $R^i$ is the next input sequence. In every timestamp $i$ we compare AHTM's predicted value to the next sequence to compare the similarity.

    RM and SM were compared on their ability to detect anomalies, with RM consistently achieving higher precision and recall.
    
    \item \textbf{Processing Speed:} The average processing time (in seconds) required to analyze each dataset was recorded. AHTM's software-based RM showed improved processing efficiency relative to HTM's SM, while H-AHTM's CAM-accelerated RM achieved even faster processing times. Performance measurements were averaged over ten trials to ensure statistical reliability~\cite{james2017htm,sp_htm}.
    
    \item \textbf{Accuracy vs. Speed Trade-off:} This metric evaluates how computational constraints impact detection accuracy and processing efficiency, which is critical for real-time anomaly detection scenarios.
\end{itemize}

Additionally, online learning capabilities were verified by continuously introducing new data streams and monitoring the system's ability to adapt without forgetting previously learned patterns. For proof of concept, large-scale datasets were used while keeping the system continuously running. RM’s biologically inspired design enabled it to improve accuracy over time, underscoring its suitability for dynamic environments~\cite{hawkins2016neurons,doyon1998role}.

\section{Results and Discussion}
\label{sec:results}

Our work leverages financial datasets to evaluate the performance of the proposed AHTM and H-AHTM. Given the nature of financial data, which is characterized by high volatility and time-sensitive decision-making, the ability to process and analyze this data swiftly is paramount. Rapid identification of anomalies and trends can offer significant advantages in financial forecasting, risk management, and trading strategies. Financial datasets demand high-performance systems due to the real-time implications of market changes. A delay in detecting anomalies or processing trends can lead to missed opportunities or financial losses. 

The results presented in the next subsections are based on the evaluation criteria discussed in Sec.~\ref{sec:Method}.

\subsection{Anomaly Detection Accuracy}

Anomaly detection accuracy measures the system's ability to identify unusual patterns or deviations in financial data, which may indicate significant market events or irregularities. This is especially important in the financial domain, where timely identification of anomalies can lead to better decision-making, risk mitigation, and improved market strategies.

Since HTM is best suited for temporal datasets, we employ a self-supervised evaluation approach, where we compare the predicted outcome at a given timestamp against the input at the next timestamp. First, anomalies are calculated based on the deviations of the predicted SDRs from its ground truth. This is done by evaluating the intersection of both matrices and calculating the inverse of an overlap score. As a result, we obtain an anomaly score, where values closer to zero represent more identical matrices. We are able to convert these scores into a more digestible accuracy score for evaluation. To determine a match between a predicted and ground truth SDRs, an intersection threshold of 0.5, or at least 50\%, must overlap. Given the nature of randomly distributed SDRs, it is unlikely for differently encoded representations of inputs to match incorrectly, ensuring the robustness of our anomaly detection and accuracy evaluation.

\begin{table*}[t]
\centering
\caption{Anomaly Detection Metrics for Financial Datasets Across HTM and AHTM}
\label{tab:anomaly_metrics_comparison}
\resizebox{\textwidth}{!}{%
\begin{tabular}{|l|c|c|c|c|c|c|c|c|}
\hline
\multirow{2}{*}{\textbf{Dataset}} & \multicolumn{4}{c|}{\textbf{HTM}} & \multicolumn{4}{c|}{\textbf{AHTM}} \\ \cline{2-9} 
                                  & Precision & Recall & F1-score & ROC-AUC    & Precision & Recall  & F1-score & ROC-AUC    \\ \hline

 Dow Jones (Weekly)                 &   0.029    &  0.032  &  0.030   &    0.516    &    0.028   &  0.031   &   0.029   &    0.515    \\ \hline
NASDAQ (Weekly)                   & 0.202     & 0.218   & 0.206     &     0.609   & 0.199      & 0.215    & 0.203     &  0.607      \\ \hline
S\&P 500 (Weekly)                 & 0.330      & 0.352   & 0.337     &   0.676     & 0.328      & 0.350    & 0.335     &   0.675     \\ \hline
Gold Prices (Monthly)              & 0.401      & 0.414   & 0.405     &    0.707    & 0.408      & 0.419    & 0.411     &    0.709    \\ \hline
S\&P 500 (Monthly)                  & 0.394     & 0.417   & 0.400     &    0.709    & 0.390      & 0.412    & 0.396     &  0.706      \\ \hline
Vix Close (Monthly)                 & 0.966      & 0.971   & 0.968     &    0.985    & 0.964      & 0.968    & 0.965     &   0.984     \\ \hline
Vix High (Monthly)                  & 0.948      & 0.952   & 0.949     &    0.976    & 0.941      & 0.946    & 0.943     &   0.973     \\ \hline
Vix Low (Monthly)                   & 0.958      & 0.963   & 0.959     &     0.981   & 0.950      & 0.955    & 0.951     &  0.978      \\ \hline
Vix Open (Monthly)                  & 0.946      & 0.951   & 0.948     &    0.976    & 0.938      & 0.942    & 0.939     &   0.971     \\ \hline

Vix Close (Daily)                 & 0.996      & 0.997   & 0.997     &     0.999   & 0.986      & 0.987    & 0.986     &     0.993   \\ \hline
Vix High (Daily)                   & 0.992      & 0.994   & 0.993     &   0.997     & 0.984      & 0.986    & 0.985     &    0.993    \\ \hline
Vix Low (Daily)	                  & 0.973      & 0.976   & 0.974     &    0.988    & 0.969      & 0.972    & 0.970     &    0.986    \\ \hline
Vix Open (Daily)                   & 0.959      & 0.962   & 0.960     &    0.981    & 0.948      & 0.951    & 0.949     &    0.975    \\ \hline
Natural Gas (Daily)               & 0.997      & 0.997   & 0.997     &    0.999    & 0.986      & 0.987    & 0.986     &    0.993    \\ \hline
Oil Prices (Daily)                  & 0.992     & 0.994   & 0.993     &   0.997     & 0.984      & 0.986    & 0.985     &   0.993     \\ \hline
\end{tabular}
} 
\end{table*}

The results in Table \ref{tab:anomaly_metrics_comparison} demonstrate that AHTM achieves comparable anomaly detection accuracy to HTM across all financial datasets while significantly reducing inference latency. The precision, recall, F1-score, and ROC-AUC values remain nearly identical between AHTM and HTM, confirming that the introduction of RM does not compromise detection reliability. This is expected, as RM is designed to accelerate first-order temporal inference rather than modify the anomaly scoring mechanism.

The key advantage of AHTM over HTM lies in its computational efficiency. Traditional HTM implementations are constrained by high inference latency and memory overhead, particularly when processing high-frequency financial data. In contrast, AHTM dramatically reduces the computational load while preserving detection performance, making it better suited for real-time anomaly detection applications where rapid response is critical.

These findings align with previous research, such as \cite{ahmad2017unsupervised}, \cite{sousa2021htm}, and \cite{lavin2015evaluating}, which demonstrated HTM’s effectiveness in anomaly detection tasks but highlighted its computational bottlenecks. AHTM addresses these bottlenecks, providing a biologically inspired, low-latency alternative for high-frequency financial data streams.

\subsection{Processing Speed}

 Processing speed is a critical metric for evaluating the real-time applicability of memory systems, particularly in financial applications. In this domain, data streams are continuous, and decisions often need to be made instantaneously. Faster processing times can lead to significant advantages in anomaly detection and predictive analysis. RM, leveraging hardware acceleration, is well-suited for such real-time tasks and offers substantial improvements in computational efficiency over SM.

Table~\ref{tab:processing_speed} presents the processing times (in milliseconds) for HTM, AHTM, and H-AHTM across various financial datasets. The table includes dataset names, temporal resolutions, and the average processing times for all three systems: HTM, AHTM, and H-AHTM. Reflex Memory incorporated HTM systems consistently demonstrate lower processing times, emphasizing its efficiency and capability to handle real-time financial data.

\begin{table}[t]
\centering
\caption{Processing Time (in milliseconds) for Financial Datasets}
\label{tab:processing_speed}
\renewcommand{\arraystretch}{1.2} 
\setlength{\tabcolsep}{3pt} 
\small 

\begin{tabular}{|c|c|c|c|}
\hline
\textbf{Dataset} & 
\textbf{\centering HTM (ms)} & 
\textbf{\centering AHTM (ms)} & 
\textbf{\centering H-AHTM (ms)} \\ \hline

S\&P 500 (Monthly) & 589.63 & 59.85 & 44.6 \\ \hline
Dow Jones (Weekly) & 729.11 & 221.87 & 209.0 \\ \hline
NASDAQ (Weekly) & 770.32 & 90.10 & 72.6 \\ \hline
S\&P 500 (Weekly) & 728.45 & 45.20 & 26.0 \\ \hline
Vix Close (Monthly) & 1189.64 & 101.80 & 61.3 \\ \hline
Vix High (Monthly) & 1309.43 & 114.57 & 72.8 \\ \hline
Vix Low (Monthly) & 1317.04 & 144.39 & 104.0 \\ \hline
Vix Open (Monthly) & 1427.79 & 131.91 & 89.8 \\ \hline
Natural Gas (Daily) & 379.10 & 262.15 & 246.0 \\ \hline
Oil Prices (Daily) & 984.29 & 196.71 & 136.0 \\ \hline
Vix Close (Daily) & 849.62 & 79.92 & 42.6 \\ \hline
Vix High (Daily) & 1349.77 & 114.88 & 71.8 \\ \hline
Vix Low (Daily) & 877.94 & 99.72 & 62.9 \\ \hline
Vix Open (Daily) & 1434.95 & 134.39 & 89.7 \\ \hline
Gold Prices (Monthly) & 233.56 & 78.58 & 73.7 \\ \hline

\end{tabular}
\end{table}

The results in Table~\ref{tab:processing_speed} highlight RM’s superior processing speed compared to SM across various financial datasets. For moderate-sized datasets, such as Monthly S\&P 500 (1,766 data points) and Monthly VIX Close (4,053 data points), AHTM achieves a notable performance improvement, processing data up to 50\% faster than HTM. This advantage becomes even more pronounced for large-scale datasets, such as Daily Natural Gas (5,802 data points) and Daily Oil Prices (8,303 data points), where RM accelerates processing by up to 70\% for our AHTM system. The scalability of AHTM is evident, as its speed advantage increases with dataset size, making it particularly effective for high-frequency financial data.

This substantial improvement in processing speed makes AHTM and H-AHTM particularly advantageous for time-sensitive applications. By reducing latency, it enables rapid anomaly detection and real-time decision-making, which are critical in the financial domain.

\begin{figure}[H]
    \centering
    \includegraphics[width=\textwidth]{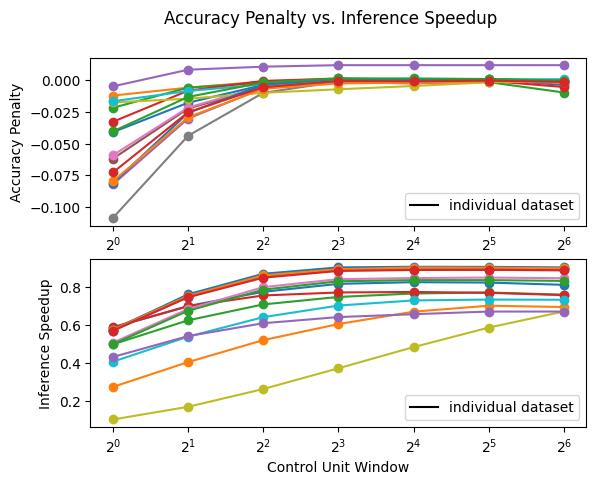}
    \caption{Impact of Increasing Control Unit Window on Accuracy and Speedup: Higher CU Windows Increase RM Reliance, Enhancing Speedup at the Cost of Slight Accuracy Reduction.}
    \label{fig:CU_window}
\end{figure}

\subsection{Accuracy vs. Speed Trade-off}
Building on the observed improvements in processing speed, further optimizations can be achieved by strategically balancing accuracy with computational efficiency. When prioritizing real-time responsiveness, slight reductions in precision can lead to substantial gains in performance. AHTM, by incorporating RM more frequently, accelerates first-order inferences while maintaining an acceptable level of predictive reliability.

The graph in Figure \ref{fig:CU_window}, illustrates the trade-off between accuracy penalty and inference speedup as a function of the CU window size. As the CU window increases, AHTM relies more on RM for inference, leading to two key effects:

\begin{itemize}
    \item \textbf{Accuracy Penalty (Top Subplot):} Initially, there is a noticeable accuracy drop as RM is used more frequently instead of Sequence Memory (SM). However, as the CU window grows, this penalty stabilizes, indicating that RM can still maintain reasonable predictive reliability even with reduced sequence depth.
    
    \item \textbf{Inference Speedup (Bottom Subplot):} The speedup effect is much more pronounced. Larger CU windows allow AHTM to bypass costly multi-order computations by leveraging RM's faster first-order inferences, leading to a significant increase in processing speed. This trend continues as the CU window expands, making inference exponentially more efficient.
\end{itemize}

Since RM excels at detecting frequently recurring patterns with low computational overhead, increasing the CU window shifts AHTM’s behavior toward prioritizing speed over precision. While SM is better suited for capturing dependencies across multiple time steps, it is computationally expensive. Thus, by relying more on RM, AHTM gains substantial speedup while sacrificing only a nominal amount of accuracy—an acceptable trade-off in many real-time applications.

This balance between efficiency and predictive reliability demonstrates that AHTM can optimize inference speed by dynamically adjusting CU window sizes. By strategically leveraging RM, AHTM achieves superior computational efficiency while keeping accuracy within tolerable limits.





       



\section{Prior Acceleration Approaches}
\label{sec:Relatedwork}

HTM has been explored extensively for unsupervised learning, anomaly detection, and sequence prediction. However, its computational overhead, particularly in SM, has led to various acceleration techniques, including hardware implementations and software optimizations. Hardware-based methods typically leverage parallel computation and specialized memory architectures, while software optimizations focus on reducing computational complexity or improving learning mechanisms.



Several hardware accelerators have been proposed for HTM. FPGA-based implementations achieve 3$\times$ speedups through parallelism but consume high power \cite{fpga_htm}. Memristor-based architectures enhance efficiency with non-volatile memory but face reliability issues \cite{memristor_htm}. Neuromorphic computing, inspired by biological learning, accelerates HTM but requires specialized hardware, limiting generalizability \cite{neuromorphic_htm}.

In contrast, software-based acceleration optimizes HTM execution in general-purpose computing. OpenCL-based HTM achieved a 632$\times$ kernel speedup and 6.5$\times$ overall acceleration via GPU parallelism \cite{opencl_htm}. Activation intensity-based learning cut training time by 29\%–61\% through adaptive learning rates \cite{activation_htm}. A GRU-enhanced HTM improved long-term prediction but increased training time by 22\%–105\%, raising computational costs \cite{gru_htm}.

Our approach introduces two key innovations that distinguish it from prior works. First, AHTM, incorporating RM, specializes in first-order inference, inspired by biological reflex arcs. Unlike previous HTM accelerations that focus on optimizing full-sequence learning, AHTM efficiently offloads repetitive patterns, reducing computational overhead while maintaining accuracy. Second, H-AHTM integrates CAM to further enhance search efficiency, enabling sub-centisecond inference speeds. Table~\ref{tab:htm_comparison} provides a structured comparison of previous HTM acceleration works with our proposed method.

\begin{table}[t]
\centering
\caption{Comparison of HTM Acceleration Methods}
\label{tab:htm_comparison}
\renewcommand{\arraystretch}{1.2} 
\setlength{\tabcolsep}{3pt} 
\small 

\begin{tabular}{|p{2cm}|p{2cm}|p{1.3cm}|p{2.5cm}|}
\hline
\textbf{Method} & 
\textbf{\centering Acceleration} & 
\textbf{\centering Speedup} & 
\textbf{\centering Limitations} \\ \hline

FPGA-HTM \cite{fpga_htm} & Hardware (FPGA) & 3$\times$ & High power usage \\ \hline
Memristor-HTM \cite{memristor_htm} & Neuromorphic Hardware & Not Reported & Device reliability challenges \\ \hline
Spin-Neuron HTM \cite{neuromorphic_htm} & Spin-Neurons + Resistive Memory & Not Reported & Requires specialized hardware \\ \hline
OpenCL-HTM \cite{opencl_htm} & GPU Software & 6.5$\times$ & High memory overhead \\ \hline

Activation Intensity-HTM \cite{activation_htm} & Self-Adaptive Learning & 1.41$\times$ to 2.56$\times$ Faster Training & Requires hyperparameter tuning \\ \hline
GRU-HTM \cite{gru_htm} & GRU-Based Learning &  22\%-105\% 
 Slower & Higher computational cost but higher accuracy in long-term predictions \\ \hline
\textbf{A-HTM} & Reflex Memory  & \textbf{7.55$\times$} & Fastest inference, biologically inspired learning \\ \hline
\textbf{H-AHTM} & Reflex Memory + CAM & \textbf{10.10$\times$} & Fastest inference, Low memory overhead \\ \hline

\end{tabular}
\end{table}


Unlike prior works, our approach modifies HTM’s algorithmic structure rather than relying solely on hardware or parallel execution. By leveraging RM, our method aims to provide scalable acceleration while maintaining HTM’s online learning capabilities.

Experimental results confirm that our AHTM model achieves a 7.55$\times$ acceleration, while H-AHTM further improves performance to a 10.10$\times$ speedup, outperforming previous HTM acceleration techniques. This demonstrates that RM and CAM integration offer a novel and more efficient pathway for accelerating HTM inference, particularly for applications requiring real-time learning.

 \section{Conclusion}
\label{sec:conclu}

This paper presents the integration of RM into neuromorphic-inspired HTM to improve processing efficiency in an Online learning model. The proposed AHTM has superior anomaly detection capabilities for first-order data, outperforming its predecessor HTM, in speed without compromising accuracy. The results highlight significant improvements in prediction times: while SM achieves an average of 0.2868 ms per prediction, it performs even better when reinforced by RM. The AHTM achieves an impressive 39.93 $\mu$s prediction time, while the hardware-accelerated CAM module H-AHTM reduces this further to just 2.65 ns.

Moreover, the integration of RM promotes online learning by enabling efficient adaptation to new patterns without retraining from scratch or exhausting memory. This capability positions the AHTM and H-AHTM as a powerful solution for real-time, adaptive systems. These findings demonstrate the potential of RM to drive accurate, faster, and scalable solutions for neuromorphic systems, making it a promising avenue for future research and practical applications.

In future work, an architecture leveraging multi-agent RM units running in parallel could significantly enhance processing efficiency. Each RM agent would operate independently while exchanging information, enabling faster adaptation and inference. A dedicated control unit could further streamline RM integration before the SP, optimizing automation at an earlier stage—before encoding to the neocortex. Additionally, multi-agent RM systems could facilitate more advanced learning paradigms, where each RM agent specializes in distinct patterns and collaborates to improve overall prediction accuracy. This distributed approach would also serve as a communication bridge, allowing RM agents to dynamically share insights, refine predictions, and enhance real-time decision-making.

\section*{Acknowledgment}

The authors would like to thank Ilia Batista for his early contributions and insightful feedback, which significantly enhanced the quality and direction of this work.

\FloatBarrier

\bibliographystyle{elsarticle-num.bst} 
\bibliography{Bib_files/HTMReflexCAM}

\begin{thebibliography}{10}
\expandafter\ifx\csname url\endcsname\relax
  \def\url#1{\texttt{#1}}\fi
\expandafter\ifx\csname urlprefix\endcsname\relax\def\urlprefix{URL }\fi
\expandafter\ifx\csname href\endcsname\relax
  \def\href#1#2{#2} \def\path#1{#1}\fi

\bibitem{bautista2020matlabhtm}
I.~Bautista, S.~Sarkar, S.~Bhanja, Matlabhtm: A sequence memory model of neocortical layers for anomaly detection, SoftwareX 11 (2020) 100491.

\bibitem{laird2017standard}
J.~E. Laird, C.~Lebiere, P.~S. Rosenbloom, A standard model of the mind: Toward a common computational framework across artificial intelligence, cognitive science, neuroscience, and robotics, Ai Magazine 38~(4) (2017) 13--26.

\bibitem{hawkins2016neurons}
J.~Hawkins, S.~Ahmad, \href{http://journal.frontiersin.org/article/10.3389/fncir.2016.00023/full}{Why neurons have thousands of synapses, a theory of sequence memory in neocortex}, Frontiers in neural circuits 10 (2016).
\newline\urlprefix\url{http://journal.frontiersin.org/article/10.3389/fncir.2016.00023/full}

\bibitem{ahmad2016neurons}
S.~Ahmad, J.~Hawkins, How do neurons operate on sparse distributed representations? a mathematical theory of sparsity, neurons and active dendrites, arXiv preprint arXiv:1601.00720 (2016).

\bibitem{lavin2015evaluating}
A.~Lavin, S.~Ahmad, Evaluating real-time anomaly detection algorithms--the numenta anomaly benchmark, in: Machine Learning and Applications (ICMLA), 2015 IEEE 14th International Conference on, IEEE, 2015, pp. 38--44.

\bibitem{zhou2018hierarchical}
T.~Zhou, Z.-Z. Zhang, Y.-Y. Chen, Hierarchical temporal memory network for medical image processing, DEStech Transactions on Computer Science and Engineering~(cmsms) (2018).

\bibitem{adam2018wafer}
K.~Adam, K.~Smagulova, O.~Krestinskaya, A.~P. James, Wafer quality inspection using memristive lstm, ann, dnn and htm, in: 2018 IEEE Electrical Design of Advanced Packaging and Systems Symposium (EDAPS), IEEE, 2018, pp. 1--3.

\bibitem{james2017htm}
A.~P. James, I.~Fedorova, T.~Ibrayev, D.~Kudithipudi, Htm spatial pooler with memristor crossbar circuits for sparse biometric recognition, IEEE transactions on biomedical circuits and systems 11~(3) (2017) 640--651.

\bibitem{neubert2018sequence}
P.~Neubert, S.~Ahmad, P.~Protzel, A sequence-based neuronal model for mobile robot localization, in: Joint German/Austrian Conference on Artificial Intelligence, Springer, 2018, pp. 117--130.

\bibitem{micheletto2018using}
R.~Micheletto, K.~Takahashi, A.~Kim, Using a hierarchical temporal memory cortical algorithm to detect seismic signals in noise, in: Science and Information Conference, Springer, 2018, pp. 855--863.

\bibitem{osegi2018using}
E.~Osegi, Using the hierarchical temporal memory spatial pooler for short-term forecasting of electrical load time series, Applied Computing and Informatics (2018).

\bibitem{zyarah2020end}
A.~M. Zyarah, K.~Gomez, D.~Kudithipudi, End-to-end memristive htm system for pattern recognition and sequence prediction, arXiv preprint arXiv:2006.11958 (2020).

\bibitem{harris2015neocortical}
K.~D. Harris, G.~M. Shepherd, The neocortical circuit: themes and variations, Nature neuroscience 18~(2) (2015) 170.

\bibitem{krestinskaya2018hierarchical}
O.~Krestinskaya, I.~Dolzhikova, A.~P. James, Hierarchical temporal memory using memristor networks: A survey, arXiv preprint arXiv:1805.02921 (2018).

\bibitem{cui2017htm}
Y.~Cui, S.~Ahmad, J.~Hawkins, The htm spatial pooler—a neocortical algorithm for online sparse distributed coding, Frontiers in computational neuroscience 11 (2017) 111.

\bibitem{zhu2015segdeepm}
Y.~Zhu, R.~Urtasun, R.~Salakhutdinov, S.~Fidler, segdeepm: Exploiting segmentation and context in deep neural networks for object detection, in: Proceedings of the IEEE Conference on Computer Vision and Pattern Recognition, 2015, pp. 4703--4711.

\bibitem{graybiel1998basal}
A.~M. Graybiel, The basal ganglia and chunking of action repertoires, Neurobiology of Learning and Memory 70~(1-2) (1998) 119--136.
\newblock \href {https://doi.org/10.1006/nlme.1998.3843} {\path{doi:10.1006/nlme.1998.3843}}.

\bibitem{seger2011habit}
C.~A. Seger, B.~J. Spiering, A critical review of habit learning and the basal ganglia, Frontiers in Systems Neuroscience 5 (2011) 66.
\newblock \href {https://doi.org/10.3389/fnsys.2011.00066} {\path{doi:10.3389/fnsys.2011.00066}}.

\bibitem{wikipedia2024procedural}
{Wikipedia Contributors}, \href{\url{https://en.wikipedia.org/wiki/Procedural_memory}}{Procedural memory}, Wikipedia, The Free EncyclopediaAccessed: 2024-03-07 (2024).
\newline\urlprefix\url{\url{https://en.wikipedia.org/wiki/Procedural_memory}}

\bibitem{doyon1998role}
J.~Doyon, R.~Laforce~Jr, G.~Bouchard, D.~Gaudreau, J.~Roy, M.~Poirier, P.~J. B{\'E}dard, F.~B{\'E}dard, J.-P. Bouchard, Role of the striatum, cerebellum and frontal lobes in the automatization of a repeated visuomotor sequence of movements, Neuropsychologia 36~(7) (1998) 625--641.

\bibitem{stevens2012role}
C.~Stevens, D.~Bavelier, The role of selective attention on academic foundations: A cognitive neuroscience perspective, Developmental cognitive neuroscience 2 (2012) S30--S48.

\bibitem{camos2018not}
V.~Camos, Do not forget memory to understand mathematical cognition, in: Heterogeneity of Function in Numerical Cognition, Elsevier, 2018, pp. 433--447.

\bibitem{malenka1999long}
R.~C. Malenka, R.~A. Nicoll, Long-term potentiation--a decade of progress?, Science 285~(5435) (1999) 1870--1874.

\bibitem{purdy2016encoding}
S.~Purdy, Encoding data for htm systems, arXiv preprint arXiv: 1602.05925 (2016).

\bibitem{cui2016continuous}
Y.~Cui, S.~Ahmad, J.~Hawkins, Continuous online sequence learning with an unsupervised neural network model, Neural computation 28~(11) (2016) 2474--2504.

\bibitem{narla22}
S.~Narla, P.~Kumar, A.~F. Laguna, D.~Reis, X.~S. Hu, M.~Niemier, A.~Naeemi, Modeling and design for magnetoelectric ternary content addressable memory (tcam), IEEE Journal on Exploratory Solid-State Computational Devices and Circuits 8~(1) (2022) 44--52.
\newblock \href {https://doi.org/10.1109/JXCDC.2022.3181925} {\path{doi:10.1109/JXCDC.2022.3181925}}.

\bibitem{moon2024afecam}
S.~H. Moon, D.~Reis, Afecam: An energy efficient analog 1fefet content addressable memory, in: Proceedings of the Great Lakes Symposium on VLSI 2024, 2024, pp. 541--545.

\bibitem{yang2019ternary}
R.~Yang, H.~Li, K.~K. Smithe, T.~R. Kim, K.~Okabe, E.~Pop, J.~A. Fan, H.-S.~P. Wong, Ternary content-addressable memory with mos 2 transistors for massively parallel data search, Nature Electronics 2~(3) (2019) 108--114.

\bibitem{yin2019_tcam}
X.~Yin, K.~Ni, D.~Reis, S.~Datta, M.~Niemier, X.~S. Hu, An ultra-dense 2fefet tcam design based on a multi-domain fefet model, IEEE Transactions on Circuits and Systems II: Express Briefs 66~(9) (2019) 1577--1581.
\newblock \href {https://doi.org/10.1109/TCSII.2018.2889225} {\path{doi:10.1109/TCSII.2018.2889225}}.

\bibitem{oyigeya2021reflex}
P.~O. Oyigeya, \href{https://ejnpn.springeropen.com/articles/10.1186/s41983-021-00307-2}{Reflex memory: A variant of implicit memory encoded interoceptively}, The Egyptian Journal of Neurology, Psychiatry and Neurosurgery 57~(1) (2021) 1--5.
\newblock \href {https://doi.org/10.1186/s41983-021-00307-2} {\path{doi:10.1186/s41983-021-00307-2}}.
\newline\urlprefix\url{https://ejnpn.springeropen.com/articles/10.1186/s41983-021-00307-2}

\bibitem{pavlov1927conditioned}
I.~P. Pavlov, Conditioned Reflexes: An Investigation of the Physiological Activity of the Cerebral Cortex, Oxford University Press, London, 1927, translated by G. V. Anrep.

\bibitem{beritashvili1974image}
I.~S. Beritashvili, Image-Driven Behavior in Animals, Nauka, Moscow, 1974, translated from Russian.

\bibitem{gupta2020temperature}
A.~Gupta, K.~Ni, O.~Prakash, et~al., Temperature dependence and temperature-aware sensing in ferroelectric fet, in: 2020 IEEE International Reliability Physics Symposium (IRPS), IEEE, 2020, pp. 1--5.

\bibitem{knudsen2008nangate}
J.~Knudsen, Nangate 45nm open cell library, CDNLive, EMEA (2008).

\bibitem{sp_htm}
T.~Ibrayev, O.~Krestinskaya, A.~P. James, Design and implication of a rule based weight sparsity module in htm spatial pooler, in: Electronics, Circuits and Systems (ICECS), 2017 24th IEEE International Conference on, IEEE, 2017, pp. 274--277.

\bibitem{ahmad2017unsupervised}
S.~Ahmad, A.~Lavin, S.~Purdy, Z.~Agha, Unsupervised real-time anomaly detection for streaming data, Neurocomputing 262 (2017) 134--147.

\bibitem{sousa2021htm}
F.~Sousa, S.~Faria, L.~Santos, R.~Mendes, Hierarchical temporal memory theory approach to stock market time series forecasting, Electronics 10~(14) (2021) 1630.
\newblock \href {https://doi.org/10.3390/electronics10141630} {\path{doi:10.3390/electronics10141630}}.

\bibitem{fpga_htm}
K.~L. Rice, T.~M. Taha, C.~N. Vutsinas, Hardware acceleration of image recognition through a visual cortex model, Optics \& Laser Technology 40~(6) (2008) 795--802.

\bibitem{memristor_htm}
D.~Fan, M.~Sharad, A.~Sengupta, K.~Roy, Hierarchical temporal memory based on spin-neurons and resistive memory for energy-efficient brain-inspired computing, IEEE Transactions on Neural Networks and Learning Systems 27~(9) (2015) 1907--1919.
\newblock \href {https://doi.org/10.1109/TNNLS.2015.2467205} {\path{doi:10.1109/TNNLS.2015.2467205}}.

\bibitem{neuromorphic_htm}
A.~M. Zyarah, D.~Kudithipudi, Neuromemristive architecture of htm with on-device learning and neurogenesis, ACM Journal on Emerging Technologies in Computing Systems 15~(3) (2019) 1--24.
\newblock \href {https://doi.org/10.1145/3344382} {\path{doi:10.1145/3344382}}.

\bibitem{opencl_htm}
M.~Lee, W.~Zhang, Optimizing hierarchical temporal memory for real-time processing with opencl, in: Proceedings of the ACM International Conference on Machine Learning, 2020, pp. 567--576.
\newblock \href {https://doi.org/10.1145/1234567.1234578} {\path{doi:10.1145/1234567.1234578}}.

\bibitem{activation_htm}
D.~Niu, L.~Yang, T.~Cai, L.~Li, X.~Wu, Z.~Wang, A new hierarchical temporal memory algorithm based on activation intensity, Computational Intelligence and Neuroscience 2022 (2022) 1--17.
\newblock \href {https://doi.org/10.1155/2022/6072316} {\path{doi:10.1155/2022/6072316}}.

\bibitem{gru_htm}
T.~Qin, R.~Chen, R.~Qin, Y.~Yu, Improved hierarchical temporal memory for online prediction of ocean time series data, Journal of Marine Science and Engineering 12~(4) (2024) 574.
\newblock \href {https://doi.org/10.3390/jmse12040574} {\path{doi:10.3390/jmse12040574}}.

\end{thebibliography}

\end{document}